%% file: 0_main.tex
\documentclass{article} 
\pdfoutput=1
\usepackage{iclr2024_conference,times}
\iclrfinalcopy
\input{math_commands.tex}

\usepackage{url}
\usepackage{hyperref}

\usepackage[utf8]{inputenc} 
\usepackage[T1]{fontenc}    
\usepackage{booktabs}       
\usepackage{amsfonts}       
\usepackage{nicefrac}       
\usepackage{xcolor}         
\usepackage{microtype}      
\usepackage[ruled]{algorithm2e}
\usepackage{amsmath}
\usepackage{caption}
\usepackage{subcaption}
\usepackage{graphicx}
\usepackage{float}
\usepackage{wrapfig}
\usepackage{multirow}
\usepackage{tablefootnote}
\usepackage[symbol]{footmisc}
\usepackage{array, makecell} %
\usepackage[nolist,nohyperlinks]{acronym}
\patchcmd{\footref}{\ref}{\ref*}{}{}
\usepackage{bbm}

\title{Intelligent Switching in Reset-Free RL}

\author{%
  Darshan Patil \thanks{Correspondence to darshan.patil@mila.quebec}\\
  Mila, Université de Montréal
  \And
  Janarthanan Rajendran \thanks{Work done during Postdoc at Mila, Université de Montréal} \\
  Dalhousie University
  \AND
  Glen Berseth \\
  Mila, Université de Montréal\\
  Canada CIFAR AI Chair
  \And
  Sarath Chandar \\
  Mila, École Polytechnique de Montréal\\
  Canada CIFAR AI Chair
}
\newcommand*{\methodfull}{Reset Free RL with Intelligently Switching Controller}
\newcommand*{\method}{RISC}

\newcommand*{\M}{\mathcal{M}}
\renewcommand*{\S}{\mathcal{S}}
\newcommand*{\A}{\mathcal{A}}
\newcommand*{\G}{\mathcal{G}}
\newcommand*{\T}{\mathcal{T}}
\newcommand*{\tabletop}{\emph{Tabletop Manipulation}}
\newcommand*{\sawyerdoor}{\emph{Sawyer Door}}
\newcommand*{\sawyerpeg}{\emph{Sawyer Peg}}
\newcommand*{\minitaur}{\emph{Minitaur}}
\newcommand*{\rooms}{\emph{4 rooms}}

\newcommand*{\new}[1]{#1}

\begin{document}

\maketitle

\begin{abstract}
  In the real world, the strong episode resetting mechanisms that are needed
  to train agents in simulation are unavailable.
  The \textit{resetting} assumption limits the potential of reinforcement
  learning in the real world, as
  providing resets to an agent usually requires the creation of additional
  handcrafted mechanisms or human interventions.
  Recent work aims to train agents (\textit{forward}) with learned resets by
  constructing a second (\textit{backward}) agent that returns the forward
  agent to the initial state.
  We find that the termination and timing of the transitions between these two
  agents are crucial for algorithm success.
  With this in mind, we create a new algorithm, \methodfull\ (\method{}) which
  intelligently switches between the two agents based on the agent's confidence
  in achieving its current goal.
  Our new method achieves state-of-the-art performance on several challenging
  environments for reset-free RL. \footnote{Code available at \url{https://github.com/chandar-lab/RISC}.}
\end{abstract}

\input{sections/1_intro}

\input{sections/2_related_work}
\input{sections/3_prelim}

\input{sections/4_theory}
\input{sections/5_method}
\input{sections/6_experiments}

\input{sections/7_conclusion}
\section*{Acknowledgements}
Sarath Chandar is supported by the Canada CIFAR AI Chairs program, the Canada Research Chair in Lifelong Machine Learning, and the NSERC Discovery Grant. Glen Berseth acknowledges funding support from the Canada CIFAR AI Chairs program and NSERC. Janarthanan Rajendran acknowledges the support of the IVADO postdoctoral fellowship. The authors would like to thank Nishanth Anand, Mohammad Reza Samsami, and anonymous reviewers for their helpful feedback and discussions. We would also like to acknowledge the material support of the Digital Research Alliance of Canada (alliancecan.ca), Mila IDT (mila.quebec), and NVidia in the form of computational resources.


\clearpage
\bibliographystyle{iclr2024_conference}
\bibliography{ref}

\clearpage

\input{1_supplementary.tex}

\end{document}

%% file: math_commands.tex
\usepackage{amsmath,amsfonts,bm}

\def\eqref#1{equation~\ref{#1}}

\def\1{\bm{1}}

\DeclareMathAlphabet{\mathsfit}{\encodingdefault}{\sfdefault}{m}{sl}
\SetMathAlphabet{\mathsfit}{bold}{\encodingdefault}{\sfdefault}{bx}{n}

\newcommand{\E}{\mathbb{E}}



%% file: sections/1_intro.tex
\section{Introduction}
Despite one of reinforcement learning's original purposes being as a way to
emulate animal and human learning from interactions with the real
world \citep{suttonReinforcementLearningIntroduction2018}, most of its recent
successes have been limited to simulation
\citep{mnihHumanlevelControlDeep2015,silverMasteringGameGo2016,adaptiveagentteamHumanTimescaleAdaptationOpenEnded2023}.
One of the reasons for this is that most work
trains agents in episodic environments where the agent is frequently and
automatically reset to a starting state for the task.
Environment resetting, while simple to do in a simulator, becomes much more
expensive, time consuming, and difficult to scale for real world applications
such as robotics, as it requires manual human intervention or the use of highly
specialized scripts.

Our current algorithms are designed around the ability to reset the environment, and do
not transfer to settings without resets~\citep{co-reyesEcologicalReinforcementLearning2020}.
Resets allow the agent to practice a task from the same initial states,
without needing to learn how to get to those initial states. This revisiting is critical
for RL agents that learn through trial and error, trying different actions from the same
states to learn which are better.
Resets also allow the agent to
automatically exit problematic regions of the state space.
It is often much
easier to enter certain regions of the state space than it is to exit them (e.g., falling
down is easier than getting back up).

Recent works have started to explore learning in environments where automatic resets are not
available in a setting known as reset-free or autonomous RL. A common approach for such settings is to have the agent switch between a forward controller that tries to learn the
task, and a reset controller that learns to reset the agent to favorable states that the
forward controller can learn from
\citep{eysenbachLeaveNoTrace2017,hanLearningCompoundMultistep2015,
    zhuIngredientsRealWorldRobotic2020,sharmaAutonomousReinforcementLearning2021a,
    sharmaStateDistributionMatchingApproach2022}.
Crucially, the part of these algorithms that switches between the controllers has gone
understudied.


This paper explores how we can improve performance by switching between
controllers more intelligently.
Prior works have not established a consistent strategy on \emph{how} to
bootstrap
the value of last state before the agent switches controllers.
Bootstrapping is when the agent
updates some value estimate using an existing value estimate of a successor state
\citep{suttonReinforcementLearningIntroduction2018}. It is a common idea used across
many RL algorithms.
\citet{pardoTimeLimitsReinforcement2022} empirically
established that different bootstrapping strategies for the last state in a trajectory
can result in different policies. We extend this result by analyzing why
different bootstrapping strategies can result in different optimal policies.
We show that bootstrapping the last state in the trajectory is crucial for reset-free RL
agents to maintain consistent learning targets, and doing so greatly improves their
performance.

Another underexplored aspect of the switching mechanism is learning \emph{when} to
switch.
Because reset-free RL is a unique setting in that there is no episode time limit imposed
by the environment while training, the duration of the agent's controllers' trajectories
becomes a parameter of the solution method. Prior work generally uses a fixed time
limit, but can learning when to switch also help the agent learn? For example, if the
agent enters a part of the state
space where it already knows how to accomplish its current goal, gathering more
experience in that area is unlikely to help it further in learning the task. Switching controllers and
thus the goal of the agent
can allow the agent to learn more efficiently by gathering experience for states and
goals it has not mastered yet.

\begin{figure}
    \centering
    \includegraphics[width=\textwidth]{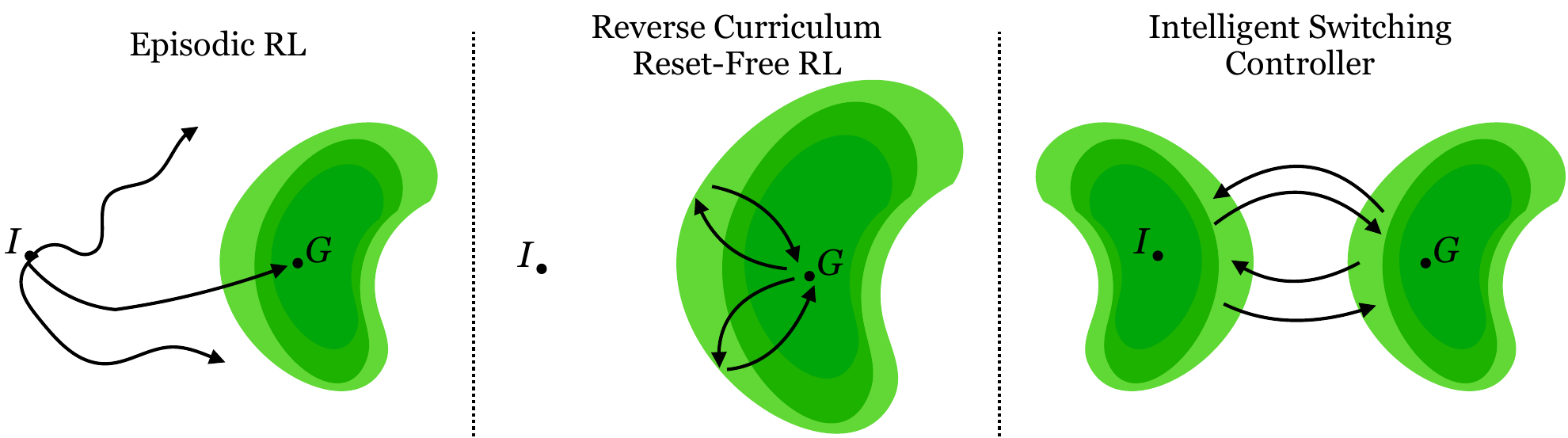}
    \caption{In RL, the agent usually starts learning about the task in areas
        around rewarding states,
        and eventually propagates the learning to other parts of the state space.
        (Left) In episodic learning, the agent starts its trajectories at a state in the initial
        state distribution. Through exploration, it might find a trajectory that
        produces a reward, but might struggle to reach that goal state again,
        particularly on sparse reward or long horizon tasks. (Center) A common approach
        in Reset-Free RL is to build a curriculum outward from the task's goal states.
        While this allows the agent to frequently visit rewarding states, it also
        means the majority of the agents experience will be generated in areas it
        has already learned. (Right) \method{} switches directions when it feels
        confident in its ability to achieve its current goal (both in the forward
        (task's goal states) or the backward direction (task's intial states)). This not
        only reduces the
        time spent in already explored regions of the state space, but also
        reduces the average distance to the goal which makes it easier for the
        agent to find high value states.}
    \label{fig:motivation}
\end{figure}

Based on these two ideas, we propose \methodfull{} (\method{}), a reset-free RL
algorithm that intelligently switches between controllers.
To reduce the amount of time spent in parts of the state space that the agent
has already learned well in, we learn a score corresponding to the agent's ability to
reach its current goal (the current goal could be either in the forward or the backward direction). The agent then switches directions with probability
proportional to that score. This allows the agent to maximize the experience
generation in areas of the state space that it still has to learn more from (Figure \ref{fig:motivation} right).
We evaluate our algorithm's performance on the recently proposed EARL benchmark
\citep{sharmaAutonomousReinforcementLearning2021}. The benchmark consists of several
robot manipulation and navigation tasks that need to be learned with minimal environment
resets, and we show that our algorithm
achieves state-of-the-art performance on several of these reset-free environments.

%% file: sections/2_related_work.tex
\section{Related Work}
\label{sec:related_work}



\label{subsec:reset_free_rl}



RL algorithms developed for episodic settings often fail on even simple tasks
when trained in reset-free environments
\citep{co-reyesEcologicalReinforcementLearning2020,sharmaAutonomousReinforcementLearning2021}.
Several settings have been proposed to address the shortcomings of episodic RL
environments
including continuing RL \citep{khetarpalContinualReinforcementLearning2022}, which tries
to maximize the training reward over the lifetime of the agent,
and single lifetime RL \citep{chenSingleLifeReinforcementLearning2022} which tries to
accomplish a task just once in a single lifetime. Other works have explored the problem
of learning to practice in separate training environments in episodic settings
\citep{rajendranHowShouldAgent2019}.
Our setting is different in that the
agent practices in a non-episodic environment, but is evaluated on a task in an episodic
environment.
Several approaches have been proposed to enable learning without resets,
including unsupervised skill learning \citep{xuContinualLearningControl2020,luResetFreeLifelongLearning2021} and
framing the problem as a multi-task learning problem
\citep{guptaResetFreeReinforcementLearning2021,guptaDemonstrationBootstrappedAutonomousPracticing2022}.

A simple approach to learn in this setting is Forward Backward RL (FBRL), which
alternates between a forward
controller that tries to accomplish the task and a reset controller that tries
to recover the initial state distribution \mbox{\citep{eysenbachLeaveNoTrace2017,hanLearningCompoundMultistep2015}}.
Other works explore different reset strategies, such as R3L where the reset
controller tries to reach novel states \citep{zhuIngredientsRealWorldRobotic2020} or MEDAL which
tries recover the distribution of states in demonstration data
\citep{sharmaStateDistributionMatchingApproach2022,sharmaSelfImprovingRobotsEndtoEnd2023}.
VapRL uses
demonstration data to make a reset controller that builds a reverse curriculum for the
forward controller
\citep{sharmaAutonomousReinforcementLearning2021a}.
\new{IBC, a concurrent work, uses optimal transport to create
    a curriculum for both the forward and reset agents without demonstrations
    \citep{kimDemonstrationfreeAutonomousReinforcement2023}.}

Early resets have been used in the context of safe reset-free RL where the reset
policy is triggered when the forward policy is about to do something unsafe
\citep{eysenbachLeaveNoTrace2017,kimAutomatingReinforcementLearning2022}. Notably, these resets happen
from the opposite direction compared to our method (see Center vs Right in
Figure \ref{fig:motivation}).
Instead of taking the agent back to areas it is confident in, as these methods
do, our method aims to maximize the amount of experience gathered in areas of
the state-goal space where the agent has not learned.

%% file: sections/3_prelim.tex
\section{Preliminaries}
\label{sec:prelim}
In this section, we define the problem of reset-free RL. Consider a goal-conditioned
Markov Decision Process (MDP): $\M \equiv (\S, \A, \G, \T, r, \rho, p_g, \gamma)$ where
$\S$ denotes the state space, $\A$ denotes the action space, $\G$ denotes the goal
space, $\T( \cdot | a, s)$ is a transition dynamics function, $r(s, a, g)$ is a reward
function based on goal $g$, $\rho$ is the initial state distribution, $p_g$ is the
desired goal distribution, and $\gamma \in [0,1]$ is a discount factor denoting
preference for long term rewards over short term rewards. In episodic RL, the objective
is to find a policy $\pi$ that maximizes
\begin{equation}
        \label{eq:rl_objective}
        J(\pi) = \E_{s_0\sim\rho, g\sim p_g,  a_t\sim \pi(s_t, g), s_{t+1}\sim
                \T(\cdot|a, s)}
        \left[\sum_t \gamma^t r(s_t, a_t, g)\right].
\end{equation}
In this formulation, sampling $s_0\sim \rho$ starts a new episode and
``resets'' the environment. Episodic RL assumes regular access to
resets, usually happening every few hundred or thousand timesteps. %

Our problem setting of reset-free RL, formalized as the ``deployment'' setting in
\citet{sharmaAutonomousReinforcementLearning2021}, involves maximizing the same
objective as above. The main difference between the episodic setting and the reset-free
setting is that the reset-free setting allows for much less frequent resetting of the
environment during training (anywhere from once every few hundred thousand steps to just once at the
beginning of training).
In the episodic case, because the agent is reset to the initial state distribution
frequently, it can directly evaluate the objective and much more easily get a signal for
how it is performing. In the reset-free setting, because the agent is reset to the
initial state distribution infrequently, the agent has to improve its performance on the
objective while getting much less direct feedback on its performance.


%% file: sections/4_theory.tex
\section{\methodfull{}}
In this section, we describe \methodfull{} (\method{}) and its components. Section
\ref{sec:theory} discusses the importance of proper bootstrapping when switching
controllers in Reset-Free RL, Section \ref{sec:esfb} describes \method{}'s early
switching mechanism, and Section \ref{subsec:alg_summary} summarizes our entire
approach.
\subsection{On Bootstrapping for Reset-Free RL}
\label{sec:theory}
In this section, we highlight the importance of proper bootstrapping when switching controllers in Reset-Free RL.
For any policy $\pi$, the value function $V^\pi(s)$ measures the expected discounted sum
of rewards when taking actions according to $\pi$ from state $s$:
\begin{equation}
    \label{eq:value_function}
    V^\pi(s) = \E_\pi\left[\sum_{t=0}^\infty\gamma^{t}R(s_{t}, a_{t}) | s_0 = s, a_t \sim \pi(\cdot | s_t)\right].
\end{equation}
For brevity and clarity of exposition, we focus our analysis on the value function, but
our analysis can be extended to state-action value functions with basic modifications. The
value function satisfies the Bellman equation
\citep{bellmanMarkovianDecisionProcess1957}:
\begin{equation}
    \label{eq:bellman}
    V^\pi(s) = \E_\pi\left[R(s, a) + \gamma V^\pi(s') | s, a \sim \pi(\cdot | s), s' \sim \T(\cdot|s, a)\right].
\end{equation}
Most RL algorithms involve learning either the value function or the state-action value
function, usually through temporal difference (TD) learning
\citep{suttonReinforcementLearningIntroduction2018}. The main mechanism of TD learning
is bootstrapping, where the value of a state is updated using the value of a successor
state with an update mirroring the Bellman equation.
When learning in an episodic environment, it is common to have time limits that end the
episode before the agent has reached a terminal state. These early resets are useful
because they can diversify the agent's experience.
The value of terminal states is generally not bootstrapped, as by definition, the value
of a terminal state is zero and it has no successors.
\citet{pardoTimeLimitsReinforcement2022} empirically showed that bootstrapping
the value of the last state in a trajectory that ended because of a timeout can result
in not only a different policy than not bootstrapping, but also better performance and
faster learning. We investigate why this is the case and discuss the implications for
reset-free RL.

Let us denote the strategy where the value function is learned by bootstrapping the last
state in a trajectory that ended in a timeout as \emph{timeout-nonterminal} and the strategy where such
states are not bootstrapped as \emph{timeout-terminal}.
Based on equation \ref{eq:bellman}, the loss for the timeout-nonterminal strategy corresponds to
\begin{equation}
    \label{eq:loss_aware}
    \mathcal{L}_n(\theta) = \E_{s\sim \xi, s'\sim P_\pi(\cdot | s)}
    \big[ \|V_\theta^{\pi}(s) - (r + \gamma V_\theta^{\pi}(s'))\| \big].
\end{equation}
Here $\xi$ corresponds to the weighting of
different states according to the experience
that is gathered by the agent, and $P_\pi(\cdot | s)$ is the transition distribution under
policy $\pi$.
The loss for the timeout-terminal strategy corresponds to
\begin{equation}
    \label{eq:loss_unaware}
    \mathcal{L}_t(\theta) = \E_{s\sim \xi, s'\sim P_\pi(\cdot | s),  d\sim \kappa}
    \big[ \|V_\theta^{\pi}(s) - (r + \gamma (1-d) V_\theta^{\pi}(s'))\| \big]
\end{equation}
The loss uses the same terms as the timeout-nonterminal strategy, except for a binary random
variable $d$ drawn from some process $\kappa$
that represents whether the episode ended because of a timeout or not.

In episodic RL, $\xi$ is the stationary distribution over states for policy $\pi$, and
depends only on $\pi$, the initial state distribution $\rho$, and the transition dynamics
function $\T$. $\kappa$ is the rule that returns 1 when the trajectory gets over a
certain length, and thus $\E[d | s]$ only depends on $\pi$, $\rho$, $\T$, and $\kappa$.
The difference in the two loss terms results in the magnitude of the value of some
states being reduced in the timeout-terminal strategy, depending on $\E[d|s]$. This
explains why different policies can emerge from the two strategies.

In reset-free RL, these parameters become more complicated. We can think of there being
two policies: the forward policy $\pi_f = \pi(\cdot | s, g_f)$ and the reset policy
$\pi_r =\pi(\cdot | s, g_r)$ where $g_f$ and $g_r$ are forward goals and reset goals
respectively. Examining the bootstrap loss for the forward policy, $\xi$ is now no
longer dependent on just $\pi_f$, $\rho$, $\T$, and forward timeout process $\kappa_f$
but also on $\pi_r$ and reset timeout process $\kappa_r$.
Furthermore, $\kappa_f$ and $\kappa_r$ are no longer static rules provided by the
environment, but are decision rules based on the agent.
Despite this, the timeout-nonterminal loss in Equation \ref{eq:loss_aware} at least maintains
the same targets for the value of each state. $\E[d|s]$, however, is now dependent on
$\pi_f$, $\pi_r$, $\rho$, $\T$, $\kappa_f$, and $\kappa_r$. This means that when using the timeout-terminal bootstrap loss (Equation \ref{eq:loss_unaware}),
\emph{the value function for the forward policy depends on the actions of the
    agent's reset policy and vice versa}.
Thus, in order to maintain consistent targets while training, \method{} uses the
timeout-nonterminal loss when switching controllers.

%% file: sections/5_method.tex
\subsection{Building a Curriculum with Intelligent Switching}
\label{sec:esfb}
Using a curriculum of tasks through the course of training has been shown to accelerate
learning in reinforcement learning
\citep{portelasAutomaticCurriculumLearning2020,florensaReverseCurriculumGeneration2017,sukhbaatarIntrinsicMotivationAutomatic2018a}.
A common approach to building these curricula is to start with tasks with initial states
near the goal, and expand outwards \citep{florensaReverseCurriculumGeneration2017},
known as a reverse curriculum.

\begin{minipage}[t][][b]{0.35\textwidth}
    \centering
    \begin{algorithm}[H]
        \small
        \caption{\methodfull{} (\method{})}\label{alg:esfb}
        \SetKwInOut{Output}{Input}
        \Output{
            Trajectory switching probability: $\zeta$
        }
        $s, g = \mathtt{env.reset}()$\\
        $t = 0$\\
        $check\_switch = \mathtt{random}() < \zeta$

        \While{True}{
            $a = \mathtt{agent.act}(s, g)$\\
            $s', r = \mathtt{env.step}(a)$\\
            $\mathtt{agent.update}(s, a, r, s', g)$\\
            $t= t +1$\\
            \If{$\mathtt{should\_switch}$(t, $\mathtt{agent}.Q_f$, s', g, check\_switch)}{
                $g = \mathtt{switch\_goals}()$\\
                $t = 0$\\
                $check\_switch = \mathtt{random}() < \zeta$
            }
            $s = s'$
        }
    \end{algorithm}
\end{minipage}
\hfill\vline\hfill
\begin{minipage}[t][][b]{0.56\textwidth}
    \begin{algorithm}[H]
        \small
        \caption{Switching function}
        \label{alg:switching}
        \KwIn{Minimum trajectory length: $m$\newline
            Maximum trajectory length: $M$\newline
            Conservative factor: $\beta$\newline
            Current state, goal: $s, g$\newline
            Length of current trajectory: $t$\newline
            Success Critic: $F$\newline
            Whether to perform switching check: $check\_switch$}
        \uIf(\tcp*[f]{Switch if reached goal}){$s == g$}{
            \Return{True}
        }
        \uElseIf(\tcp*[f]{Truncate after $M$ steps}){$t \geq M$}{
            \Return{True}
        }
        \uElseIf(\tcp*[f]{Min. Length Check}){$t < m$}{
            \Return{False}
        }
        \uElseIf{$check\_switch$}{
        $c=F_{\pi}(s, g)$ \tcp*{agent competency}
        $\lambda=c \times (1-\beta ^ t)$ \tcp*{P(switch)}
        \Return{$random() < \lambda$}
        }
        \Else{\Return{False}}
    \end{algorithm}
\end{minipage}

Some Reset-Free RL methods explicitly build a reverse curriculum
\citep{sharmaAutonomousReinforcementLearning2021a}; other methods while not building a
curriculum still follow the paradigm of always returning to the goal state.
While reverse curriculums are a popular idea, we ask if they are the best
approach for reset-free RL, as they can involve the agent spending a significant amount
of environment
interactions in parts of the state space it has already mastered (Figure
\ref{fig:motivation}, middle). Instead, we hypothesize that reset-free RL can benefit
from a new type of curriculum that works from the outside in.

Our algorithm is based on the intuition that if the agent is already competent
in an area of the state space, then it does not need to continue generating more
experience in that part of the state space. Instead, it could use those interactions to
learn in parts of the state space it still needs to learn, leading to more sample
efficient learning.
To that end, we introduce
\methodfull{} (\method{}). Similar to Forward-Backward RL
\citep{hanLearningCompoundMultistep2015,eysenbachLeaveNoTrace2017}, \method{} alternates
between going forward towards the task goal states and resetting to the task
initial states. Unlike Forward-Backward RL and other reset-free methods, \method{}
switches not only when it hits some predefined time limit or when it reaches its
current goal (in the forward or the backward direction), but
also when it is confident that it can reach its current goal. This allows it to
spend more time exploring regions of the state-goal space that it is unfamiliar
with (Figure \ref{fig:motivation}, right).


\subsubsection{Evaluating the Agent's Ability to Achieve its Goals}
\label{subsec:success_critic}
To build our intelligent switching algorithm, we need a way to estimate if an agent is
competent in an area of the state space. A natural way to do this is to incorporate some
notion of how many timesteps it takes the agent to achieve its goal from its current state.
Thus, we define the competency of a
policy $\pi$ in state $s$ on goal $g$ as
\begin{equation}
    \label{eq:success}
    F_\pi(s, g) = \gamma_{sc}^t, t=\min\{t': s_{t'} = g, s_0=s, a_t\sim\pi(s_t,g), s_{t+1}\sim \T(\cdot|a,s)\},
\end{equation}
where $\gamma_{sc} \in [0,1]$ is the success critic discount factor and $t$ represents
the minimum number of steps it would
take to get from state $s$ to goal $g$ when following policy $\pi$.
This metric has several appealing qualities. First, it is inversely correlated with how
many timesteps it takes the agent to reach its goal, incorporating the intuition that
becoming more competent means achieving goals faster. It also naturally assigns a lower
competency to states that are further away from a goal, which incorporates the intuition
that the agent should be more uncertain the further away it is from its goal. Finally, this metric
can be decomposed and computed recursively, similar to value functions in RL. This
allows us to use the same techniques to learn this metric as we use to learn value
functions.

We assume the ability to query whether a given state matches a
goal,
but \emph{only} for goals corresponding to the task goal distribution
or the initial state distribution. As the goal and initial state distributions represent
a small portion of states, learning such a function is not an unreasonable requirement
even for real-world applications.
We can then train a \emph{success critic} ($Q_F$) to estimate $F_\pi(s,g)$. Specifically, given a success
function $f: \S\times\G \rightarrow \{0, 1\}$ that determines if state $s$ matches goal $g$, we train a
network to predict
\begin{equation}
    \label{eq:succes-q}
    Q_{F}(s_t, a_t, g) = \E_{\pi}\left[f(s_{t + 1}, g) + (1 - f(s_{t+1}, g)) \gamma_{sc}
        \hat{Q}_{F}(s_{t+1}, a_{t+1}, g)\right],
\end{equation}
where $Q_{F}$ and $\hat{Q}_{F}$ are the success critic and target success critic
respectively.
The structure and training of the success critic is similar to
that of value critics used in Q-learning or actor critic
algorithms.
In fact, it is exactly
a Q-function trained with rewards and terminations coming from the success function.
The value of this success critic should
always be between 0 and 1.
At inference time, we compute the value of the success critic as
$F_{\pi}(s, g) = \E_{a\sim\pi(s, g)} [Q_{F}(s, a, g)]$.

Because we want the success critic's value to mirror the agent's ability to actually
achieve the goal, we train $Q_{F}$
on the same
batch of experience used to train the agent's policy $\pi$ and critic $Q$ and set
$\gamma_{sc}$ to the value of the agent's critic's discount factor.

\subsubsection{Modulating the Switching Behavior}
\label{subsec:modulations}
Because $F_{\pi}(s, g)\in[0,1]$, it might be tempting to use it directly a probability of
switching. It's important to note that the output of the success critic does not
actually correspond to any real probability; it is simply a metric corresponding to the
competency of the agent. Furthermore, switching with probability based solely
on the values output by success critic can lead to excessive switching. If the agent
is in an area of the state-goal space where $F_{\pi}(s, g)\approx 0.1$, the expected
number of steps the agent will take before switching goals is only 10. If the
agent is in a region with even moderately high success critic values for both forward and reset
goals, it could get stuck switching back and forth in that region and not
explore anywhere else. The switching behavior also alters the
distribution of states that the agent samples when learning, shifting it further from
the state distribution of the current policy, which can lead to unstable learning.

To counteract the excessive switching behavior, we introduce three mechanisms to
modulate the frequency of switching:
\textbf{(1) Switching on Portion of Trajectories:} Inspired by how
$\epsilon$-greedy policies only explore for a certain portion of their actions,
we only apply our intelligent switching to any given forward or backward trajectory
with probability $\zeta$. This decision is made at the start of the trajectory.
\textbf{(2) Minimum Trajectory Length:} For the trajectories we do early
switching for, we wait until the trajectory has reached a minimum length $m$
before making switching decisions.
\textbf{(3) Conservative Factor:} Finally, we define the probability of switching
as a function of $F_{\pi}(s, g)$ and trajectory length, that increases with trajectory length.
Specifically, $P(switch)= F_{\pi}(s, g) \times (1-\beta^t)$ where $t$ is the number
of timesteps in the current trajectory and $\beta \in [0,1]$ is a parameter that
decides how conservative to be with the switching. As $t\rightarrow \infty,
    P(switch) \rightarrow F_{\pi}(s, g)$.

\subsection{Algorithm Summary}
\label{subsec:alg_summary}
The early switching part of our method is summarized in Algorithms \ref{alg:switching}
and \ref{alg:esfb}.
At the start of a trajectory, our method decides whether to perform early switching checks for
the current trajectory.
Switching checks are only performed if the trajectory reaches some minimum length.
Finally, the probability of switching is computed as a function of the success
critic that increases with trajectory length. The agent samples a bernoulli random
variable with this probability to decide whether to switch goals.

We also do timeout-nonterminal bootstrapping
as described in Section \ref{sec:theory}
when training our agent's critic and
success critic.

\begin{figure}
    \centering
    \hfill
    \subcaptionbox{Tabletop Manipulation}{
        \captionsetup{format=hang}
        \includegraphics[width=.18\textwidth]{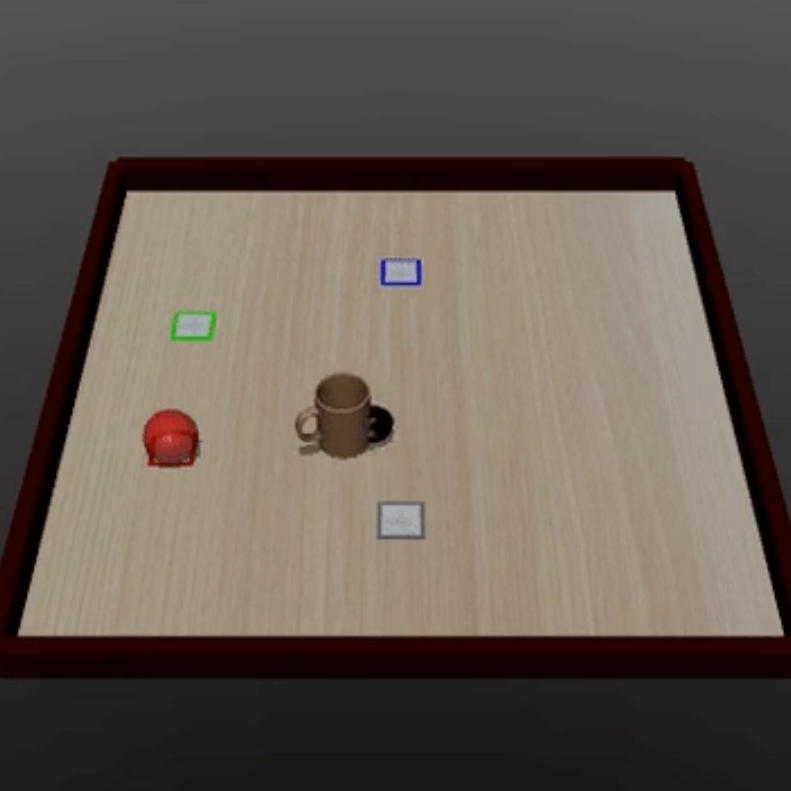}
    }\hspace{-.01\textwidth}
    \subcaptionbox{Sawyer Door}{
        \includegraphics[width=.18\textwidth]{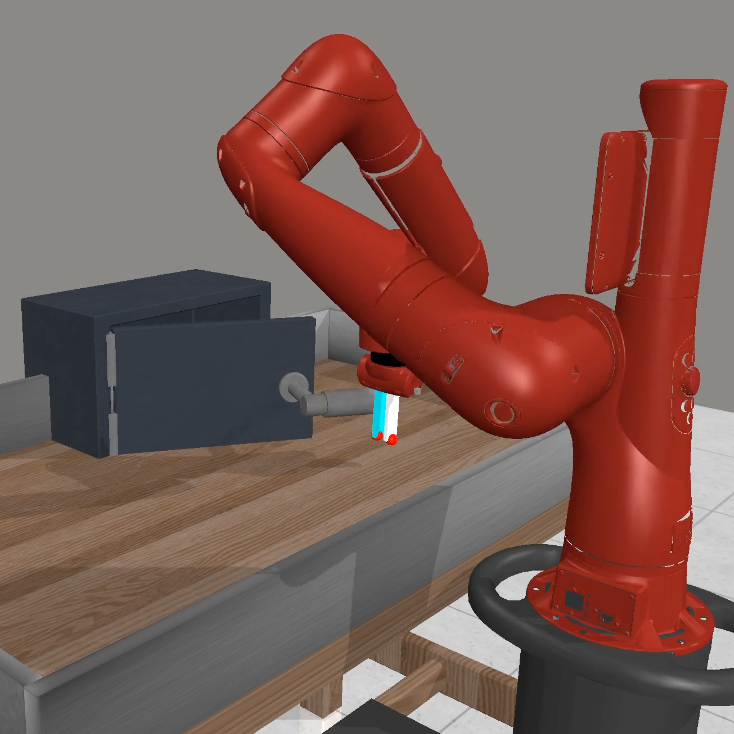}
    }\hspace{-.01\textwidth}
    \subcaptionbox{Sawyer Peg}{
        \includegraphics[width=.18\textwidth]{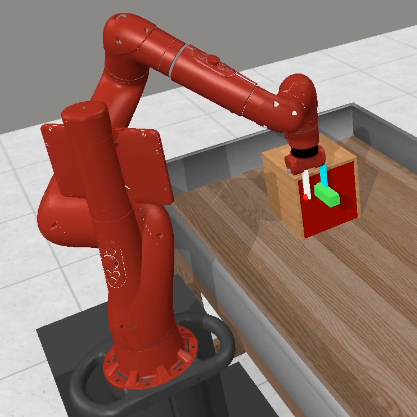}
    }\hspace{-.01\textwidth}
    \subcaptionbox{Minitaur}{
        \includegraphics[width=.18\textwidth]{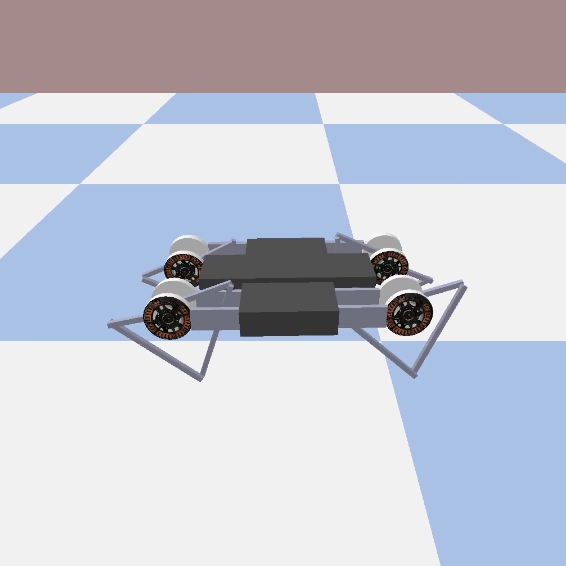}
    }\hspace{-.01\textwidth}
    \subcaptionbox{4 Rooms}{
        \includegraphics[width=.18\textwidth]{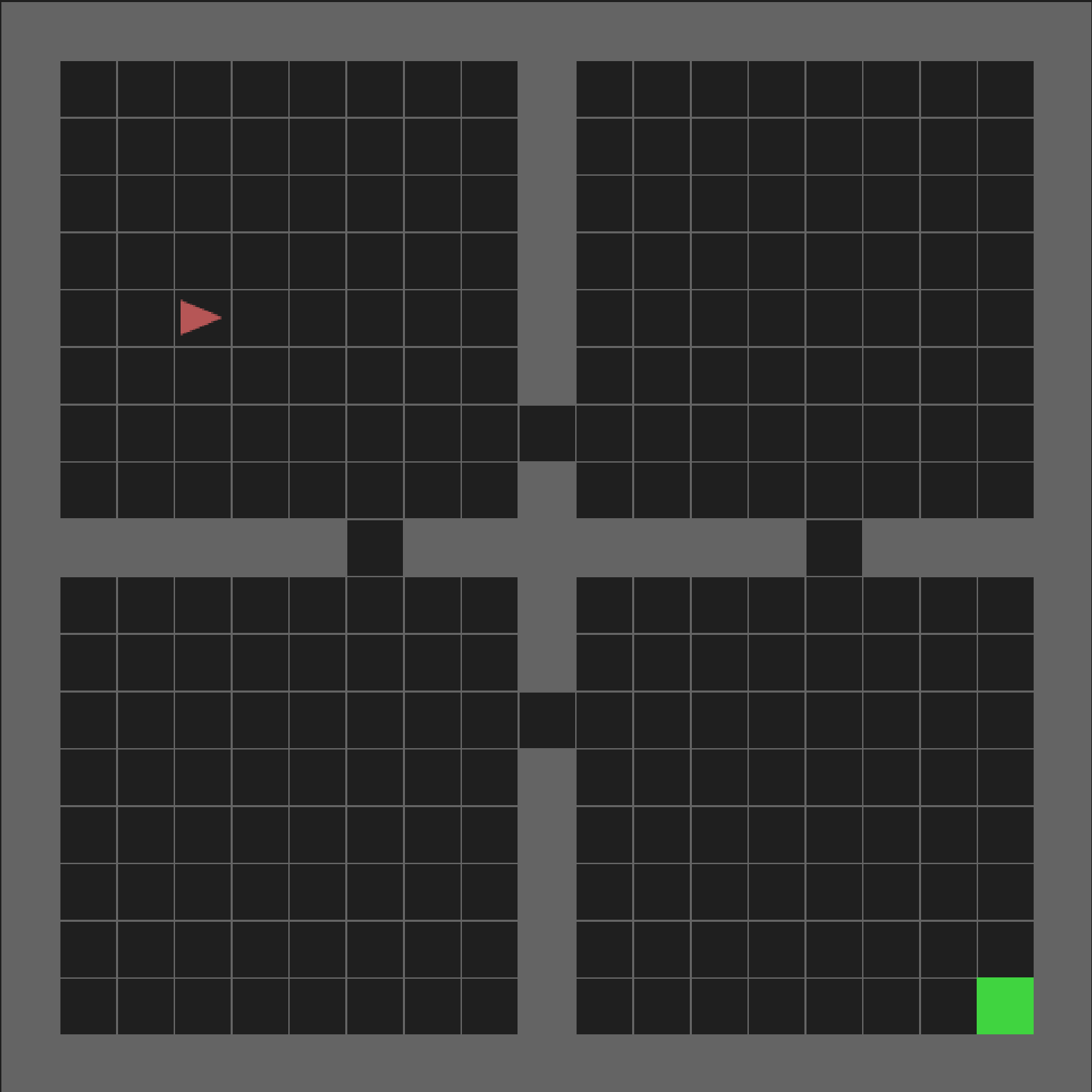}
    }\hfill
    \caption{The environments used in Section \ref{sec:experiments}. The first 4
        are part of the EARL benchmark \citep{sharmaAutonomousReinforcementLearning2021}, and the last
        is based on Minigrid \citep{chevalier-boisvertMinimalisticGridworldEnvironment2018}.}
    \label{fig:environments}
\end{figure}




%% file: sections/6_experiments.tex
\section{Experiments}
\label{sec:experiments}
In this section, we empirically analyze the performance of \method{}.
Specifically, we:
(1) Investigate whether reverse curriculums are the best approach for reset-free RL;
(2) Compare the performance of \method{} to other reset-free methods on the EARL benchmark;
(3) Evaluate the necessity of both;
timeout-nonterminal bootstrapping and early switching for \method{} with an ablation study.
\vspace{-.7cm}
\paragraph{Environments.}
The experiments in Section \ref{subsec:analysis} use a \rooms{} gridworld where the agent needs
to go from one corner to the opposite corner. The representation consists of 3
binary image channels, corresponding to the agent, the walls, and the goal location,
while each cell corresponds to a spatial location.
Sections \ref{subsec:earl_results} and \ref{subsec:ablations} use four environments from the EARL
benchmark \citep{sharmaAutonomousReinforcementLearning2021}: the \tabletop{}
environment \citep{sharmaAutonomousReinforcementLearning2021a} involves moving a mug to one of
four locations with a gripper; the \sawyerdoor{} environment
\citep{yuMetaWorldBenchmarkEvaluation2019} has a sawyer robot learn to close a door; the
\sawyerpeg{} environment \citep{yuMetaWorldBenchmarkEvaluation2019} has a sawyer robot learning to
insert a peg into a goal location; the \minitaur{} environment
\citep{coumansPyBulletPythonModule2016} is a locomotion task where a minitaur robot learns
to navigate to a set of goal locations. The first 3 environments are sparse
reward tasks where a set of forward and resetting demonstrations are provided to
the agent, while the last environment is a dense reward environment with no
demonstrations provided.

These environments all provide a low dimensional state
representation for the agent. We follow the evaluation
protocol outlined by
\citet{sharmaAutonomousReinforcementLearning2021}, evaluating the agent for 10 episodes
every 10,000 steps.
Every agent was run for 5 seeds.
For all environments, we assume access to a reward
function and a success
function that can be queried for goals in the task goal distribution or initial
\begin{wrapfigure}{r}{.48\textwidth}
  \vspace*{-.4cm}
  \centering
  \includegraphics[width=.48\textwidth]{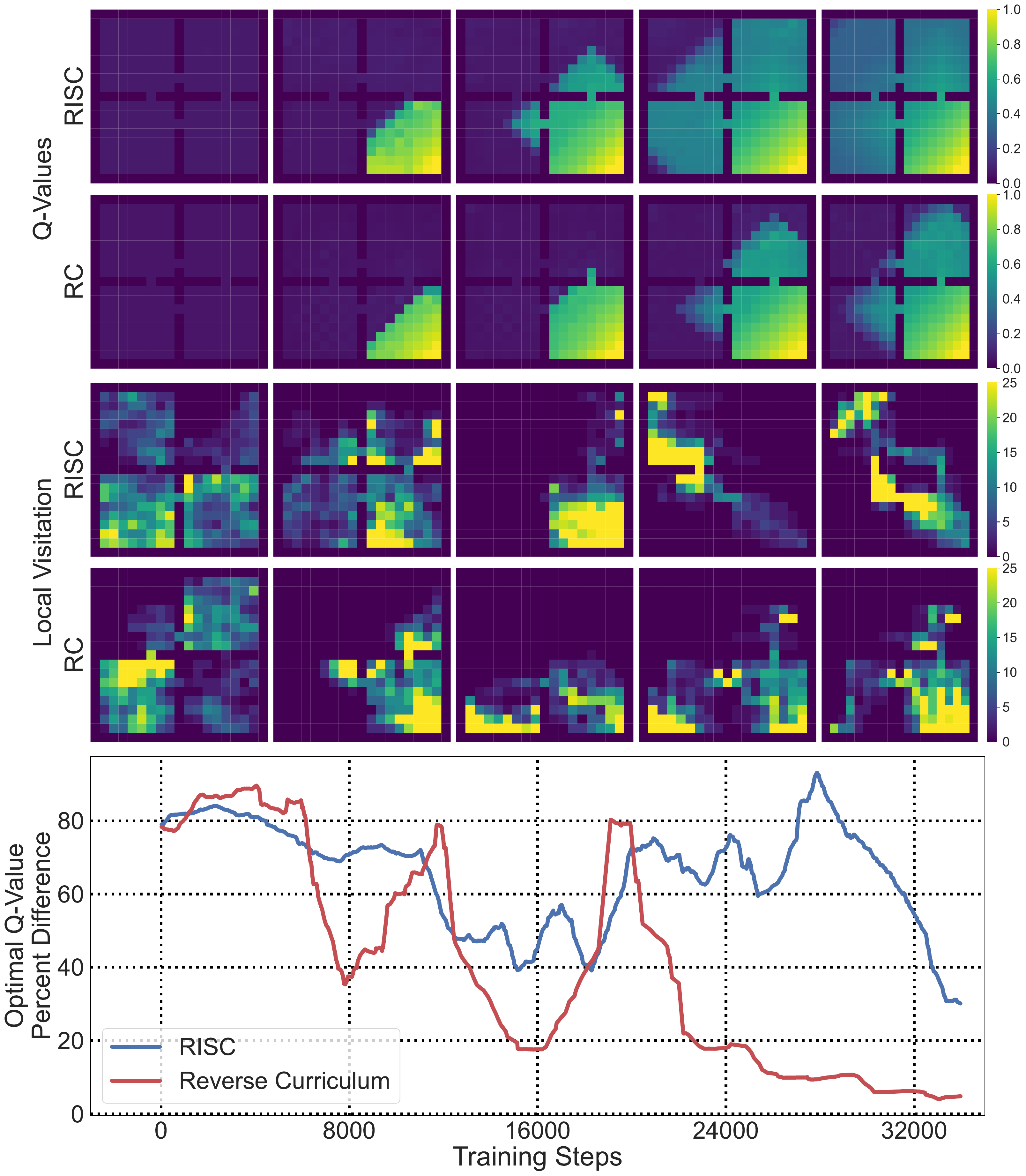}
  \caption{The heatmaps show the progression of the Q-values (top)
    and the visitation frequencies (middle) of the agent in forward mode in the $2000$ steps
    after the noted timestep.
    The reverse curriculum method  tends to
    be biased towards states that it has already learned, while \method{} is
    more evenly distributed, and slightly biased towards states it hasn't fully learned.}
  \label{fig:4rooms_analysis}
  \vspace*{-.7cm}
\end{wrapfigure}
state distribution.

\vspace{-20pt}

\paragraph{Comparisons.}

We compare against the following baselines and methods:
(1) Forward Backward RL (FBRL) \citep{eysenbachLeaveNoTrace2017,hanLearningCompoundMultistep2015},
where the agent alternates between going forward to the goal and resetting back
to the initial state distribution;
(2) R3L \citep{zhuIngredientsRealWorldRobotic2020}, which uses a novelty based reset controller;
(3) VapRL \citep{sharmaAutonomousReinforcementLearning2021a}, where the reset controller builds a
curriculum using states in demonstration data for the forward agent based on the
forward agent's performance;
(4) MEDAL \citep{sharmaStateDistributionMatchingApproach2022}, where the reset controller
learns to reset to states in the demonstration data;
(5) Naive RL, where the agent only optimizes for the task reward throughout
training i.e., an episodic agent being trained in a reset-free environment;
(6) Episodic RL, an agent run in an episodic environment with frequent automatic
resets and is meant as an unrealistic \textit{oracle} to understand the best possible performance.
The
performances for these methods were either sourced from publicly available
numbers \citep{sharmaAutonomousReinforcementLearning2021} or recreated with public implementations.

For the EARL experiments, all agents use a SAC agent \citep{haarnojaSoftActorCriticOffPolicy2018} as the base agent, with similar
hyperparameters to earlier works. For the \rooms{} experiments,
all agents use a DQN \citep{mnihHumanlevelControlDeep2015} agent as their base.

\subsection{Curricula in Reset-Free RL}
\label{subsec:analysis}

We first investigate the performance of \method{} on a \rooms{} minigrid environment,
comparing it to a simple Reverse Curriculum (RC) agent.
The RC agent
maintains a success critic similar to \method{}.
When going backward, it switches to the
forward controller if the success critic value falls below a given threshold. While this
is not the same as other reset-free reverse curriculum methods in the literature, it is simple to
implement, does not require mechanisms such as demonstrations or reward
functions for arbitrary states, and is a useful tool to see differences in behavior of
the agents.

Figure \ref{fig:4rooms_analysis} tracks the behavior and internals of an \method{} and
a RC agent across time. It shows, for the noted timesteps (1) the
Q-values of each policy (Figure \ref{fig:4rooms_analysis}, top), (2) the visitation
frequency of both agents for the 2000 steps
following the timestep (Figure \ref{fig:4rooms_analysis}, middle) (3) the difference in
the agent's state value estimate and the optimal state value estimate for the experience that
each agent gathers (Figure \ref{fig:4rooms_analysis}, bottom). We call this
the Optimal Value Percent Difference (OVPD), and it is
calculated as $OVPD(s) = \frac{|max_a Q_\theta(s, a) - V^*(s)|}{V^*(s)}$.
We use it
as a proxy for how useful the state is for the agent to collect.

The figure shows that the \method{} agent tends to gather experience in areas that it
has not yet mastered (i.e., Q-values are low), while the RC agent gathers much more
experience in areas that it has already mastered. This is supported both by comparing
the local visitation frequencies of the agents to the Q-values, and the OVPD values. The
OVPD of the RC agent drops very low even before it has converged, meaning that the RC
agent is not collecting transitions that it has inaccurate estimates for. Meanwhile, the
\method{} agent's OVPD stays relatively high throughout training, until it converges.
This analysis suggests that using a reverse curriculum might not be the best approach
for Reset-Free RL.

\subsection{Evaluating \method{} on the EARL benchmark}
\label{subsec:earl_results}
\begin{figure}
  \centering
  \includegraphics[width=.55\textwidth]{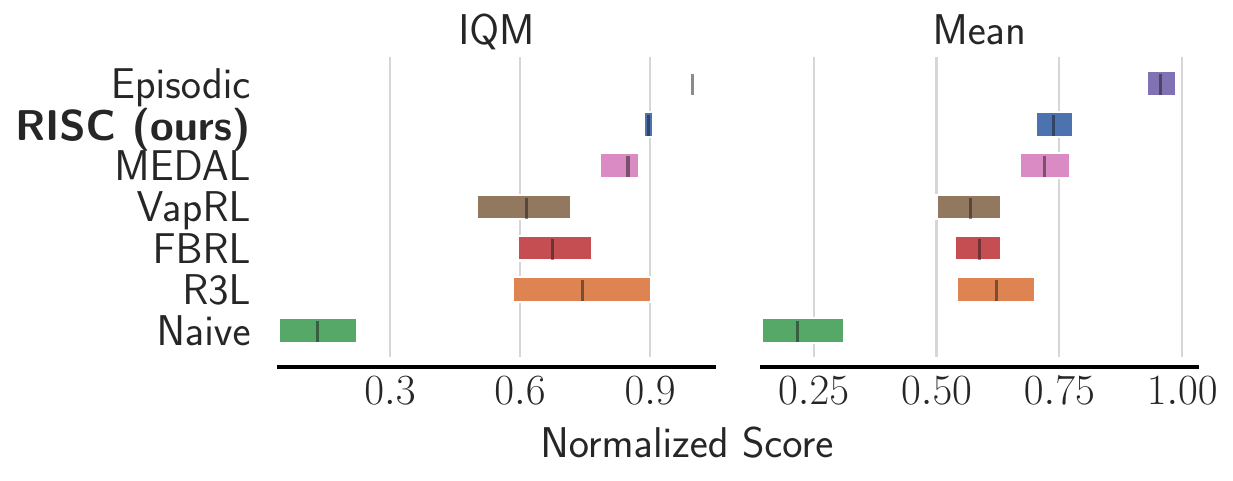}
  \includegraphics[width=.44\textwidth]{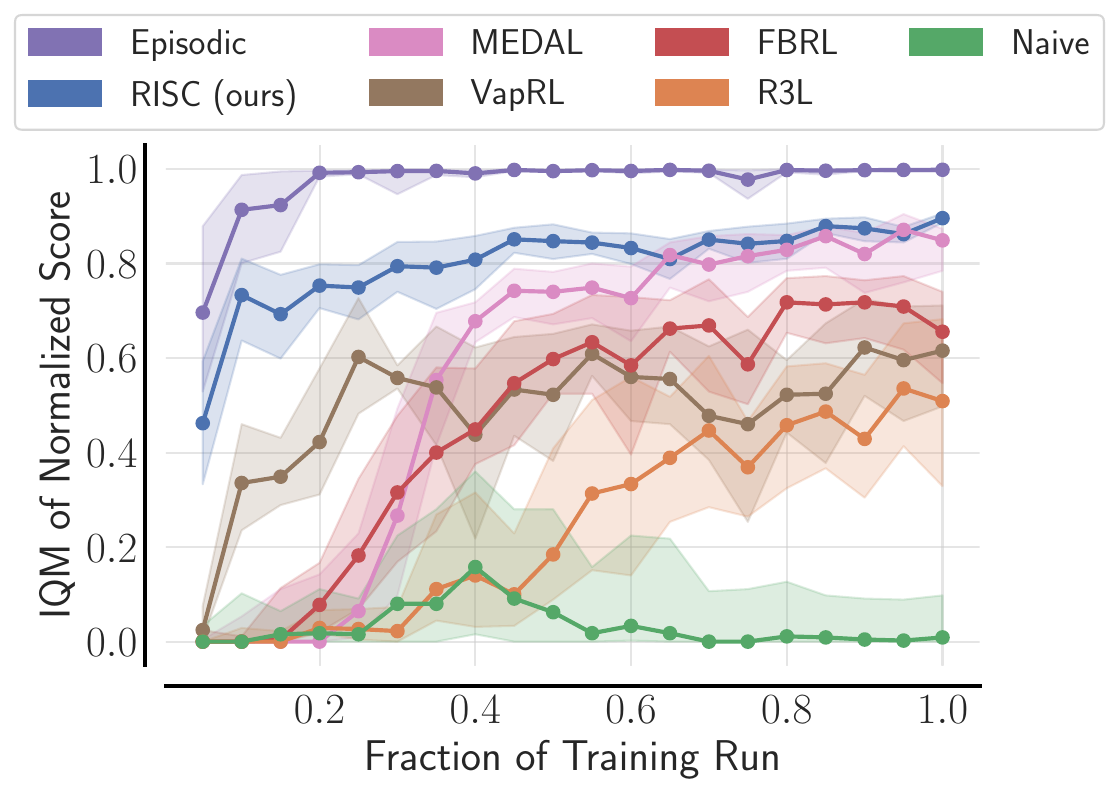}
  \caption{95\% confidence intervals for the interquartile mean (IQM) and mean
    normalized performance of \method{} and other
    baselines aggregated across environments on the EARL benchmark (\tabletop{},
    \sawyerdoor{}, \sawyerpeg{}, and \minitaur{}).
    Because MEDAL and VapRL require demonstrations and thus do not work on \minitaur{},
    we exclude \minitaur{} from their calculations (left).
    IQM of \method{} and other baselines on EARL benchmark as a function of progress
    through the training run. Shaded regions represent 95\% confidence intervals
    (right). \method{} outperforms and learns much faster than other reset-free
    baselines.
  }
  \label{fig:main_rliable}
\end{figure}
\begin{figure}[ht]
  \centering
  \includegraphics[width=\textwidth]{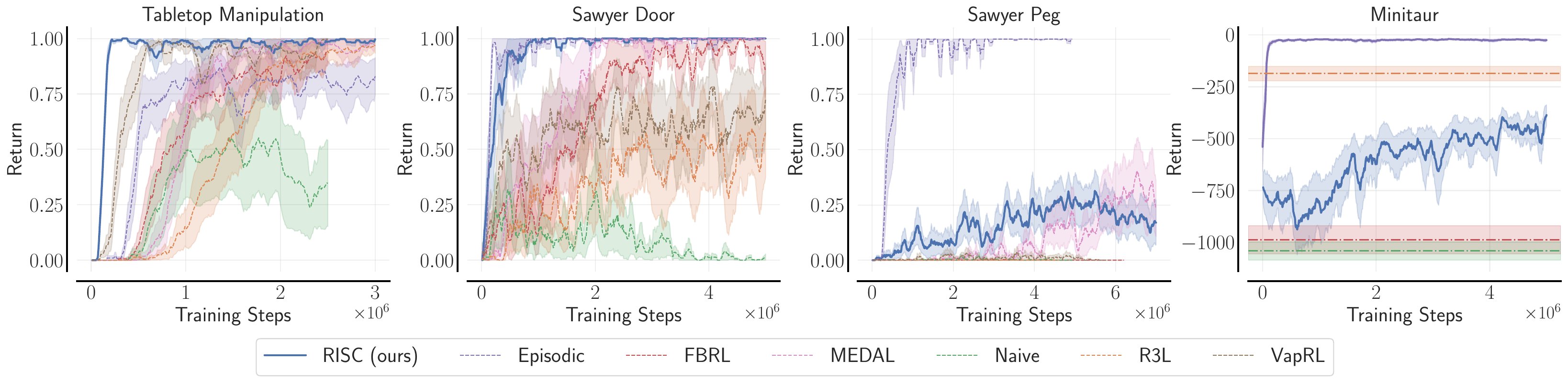}
  \caption{
    Average test returns on EARL benchmark tasks over timestep.
    \method{} improves upon or matches the state-of-the-art for reset-free algorithms on
    3 of the 4 environments (\tabletop{}, \sawyerdoor{}, \sawyerpeg{}), and even
    outperforms/matches the Episodic baseline on \sawyerdoor{} and \sawyerpeg{}.
    Learning curves and code were not available for several baselines for \minitaur,
    so only the final performance is plotted.
    Results are averaged over 5 seeds and the shaded regions represent standard
    error.
  }
  \label{fig:learning_curves}
\end{figure}
Figures \ref{fig:main_rliable} and \ref{fig:learning_curves} show the results of
\method{}
on the EARL benchmark environments compared to previously published results.
The aggregate performance of \method{} on the EARL benchmark is better than any other
reset-free method. \method{} also learns the fastest both in aggregate and across multiple
individual environment in the benchmark. In the scope of reset-free RL, it is
fairly expensive to collect data, so any method that can reach the better policy faster
is of great value.

It is also important to note the requirements and complexity of some of these methods.
Both MEDAL and VapRL require demonstrations, \emph{meaning that they cannot be run on
  environments such as Minitaur}. VapRL also requires access to a reward function that
can be queried for goal states in the demonstration data. \method{} uses demonstration
data to seed the replay buffer, but does not require demonstrations to be provided.
When comparing \method{} to the methods that can run on all environments, \method{}
clearly outperfoms them all.

\subsection{Ablating the components of \method{}}
\label{subsec:ablations}
\begin{figure}
  \includegraphics[width=\textwidth]{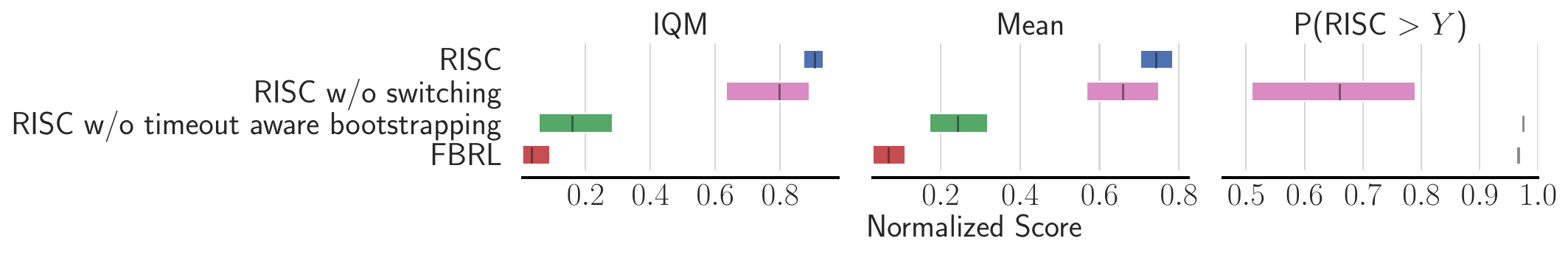}
  \caption{95\% confidence intervals for the interquartile mean (IQM), mean, and
    Probability of Improvement of \method{} for ablations of \method{}.}
  \label{fig:ablation_rliable}
\end{figure}

We now ablate the components of \method{} to better understand the contributions of
each. We rerun \method{} with early switching removed and with the timeout aware
bootstrapping removed. Removing both components results in a method that is equivalent
to FBRL, and so we also include that in our comparison. From Figure
\ref{fig:ablation_rliable}, we see that removing the timeout aware bootstrapping results
in a significant drop in performance, and removing early switching also results in a
modest drop.
This suggests that both components are important for the success of
\method{}.

These results are also consistent in the context of our analysis in Section
\ref{sec:theory}. Our arguments there imply that not doing timeout aware bootstrapping
can be a large issue for reset-free methods, and especially methods such as ours where
the trajectory lengths can be significantly shorter than the episode lengths for the
deployment task.

%% file: sections/7_conclusion.tex
\section{Conclusion and Future Work}
\label{sec:conclusion}
We propose a method that learns to intelligently switch between controllers in reset-free RL.
It does so by (1) careful management of bootstrapping of states when switching
controllers and (2) learning when to actually switch between the controllers.
Our method results in performance that
sets or matches the state-of-the-art on 3 of the 4 environments from the recently
proposed EARL benchmark. We show success across both sparse
reward tasks with demonstrations and dense reward tasks with no demonstrations.
Finally, we note that at least two of the
environments, \sawyerdoor{} and \tabletop{}, seem close to being saturated in terms of
performance, and future work should explore other environments or ways of increasing the
difficulty of those environments.


One limitation of our method is that it may not be well suited for environments with
irreversible states, where the agent could get stuck. \method{} tries to take advantage
of the reset-free RL setting by exploring the state space more aggressively. While this
is not an issue with the environments we study in this work, as they do not contain any
irreversible states, it likely could be problematic in environments where there are
irreversible states and safety is a concern. We also do not leverage demonstrations to
guide our agent intelligently as previous works do, and learning to do so could be an
interesting avenue for future work.

%% file: 1_supplementary.tex
\appendix

\section{Agent Details}
All of our agents for the experiments on the EARL benchmark
\citep{sharmaAutonomousReinforcementLearning2021}
use SAC \citep{haarnojaSoftActorCriticOffPolicy2018} as the base agent.
The neural network architecture used for the SAC agent was taken from the publicly
available implementation of the MEDAL algorithm \citep{sharmaStateDistributionMatchingApproach2022}.
A Linear Layer (50 units), LayerNorm, and Tanh activation are applied to the observation before
being fed to the actor or critic networks.
The other hyperparameters used for the base agent are described in Table \ref{tab:sac_hyperparameters}.
The experiments on the 4-rooms gridworld \citep{chevalier-boisvertMinimalisticGridworldEnvironment2018}
use DQN \citep{mnihHumanlevelControlDeep2015} as the base agent.
The corresponding hyperparameters for those experiments are shown in Tables
\ref{tab:dqn_hyperparameters}.
The additional hyperparameters for \method{} are shown in Table \ref{tab:sac_esfb_hyperparameters}.

For the environments that provide demonstrations (listed in \ref{sec:experimental_setup}), we simply
insert those demonstrations into the replay buffer of the agents at the start of training.

\begin{table}[ht]
    \begin{minipage}{.5\linewidth}
        \begin{tabular}{ll}
            \toprule
            \textbf{Hyperparameter} & \textbf{Value} \\
            \midrule
            Actor Network           & MLP(256, 256)  \\
            Critic Network          & MLP(256, 256)  \\
            \midrule
            Optimizer               & Adam           \\
            Actor LR                & 3e-4           \\
            Critic LR               & 3e-4           \\
            $\alpha$ LR             & 3e-4           \\
            \midrule
            Weight Initialization   & Xavier Uniform \\
            Bias Initialization     & Fill(0)        \\
            \midrule
            Target entropy scale    & 0.5            \\
            Reward Scale Factor     & 10             \\
            Critic Loss Weight      & 0.5            \\
            \midrule
            Target update $\tau$    & 0.005          \\
            Target update frequency & 1              \\
            \midrule
            Discount Factor         & 0.99           \\
            Batch Size              & 256            \\
            Initial Collect Steps   & 10,000         \\
            Replay Capacity         & 10,000,000     \\
            \makecell[l]{Collect Steps               \\per Update} & 1              \\
            \midrule
            Min Log Std             & -20            \\
            Max Log Std             & 10             \\
            Log Std Clamp function  & Clip           \\
            \bottomrule
        \end{tabular}
        \caption{Hyperparameters for base SAC Agent}
        \label{tab:sac_hyperparameters}
    \end{minipage}%
    \begin{minipage}{.5\linewidth}
        \begin{tabular}{ll}
            \toprule
            \textbf{Hyperparameter}       & \textbf{Value}                    \\
            \midrule
            Q-Network                     & \makecell[l]{Conv([16,16,16], 3), \\FC(64)} \\
            \midrule
            Optimizer                     & Adam                              \\
            Q-Network LR                  & 1e-3                              \\
            \midrule
            \makecell[l]{Target hard                                          \\update frequency}  & 500                               \\
            \midrule
            Discount Factor               & 0.95                              \\
            Batch Size                    & 128                               \\
            Initial Collect Steps         & 512                               \\
            Replay Capacity               & 50,000                            \\
            \makecell[l]{Collect Steps                                        \\per Update}      & 1                                 \\
            \midrule
            $\epsilon$-greedy init value  & 1.0                               \\
            $\epsilon$-greedy end value   & 0.1                               \\
            $\epsilon$-greedy decay steps & 10,000                            \\
            \bottomrule
        \end{tabular}
        \caption{Hyperparameters for base DQN Agent}
        \label{tab:dqn_hyperparameters}
    \end{minipage}%
\end{table}

\begin{table}[ht]
    \small
    \centering
    \begin{tabular}{lll}
        \toprule
                                               & \textbf{Hyperparameter}                                                                                   & \textbf{Value}                    \\
        \midrule
        \multirow{7}{*}{Tabletop Manipulation} & Conservative Factor $\beta$                                                                               & 0.9                               \\
                                               & Minimum Trajectory Length\tablefootnote[1]{As a fraction of the maximum trajectory length\label{refnote}} & 0.5                               \\
                                               & Trajectories Proportion $\zeta$                                                                           & 1.0                               \\
                                               & Success Critic Network                                                                                    & MLP(256, 256)                     \\
                                               & Success Critic Optimizer                                                                                  & Adam                              \\
                                               & Success Critic LR                                                                                         & 3e-4                              \\
                                               & Success Critic Output Activation                                                                          & $-0.5cos(x) - 1$                  \\
                                               & \# Actions Sampled                                                                                        & 5                                 \\
        \midrule
        \multirow{7}{*}{Sawyer Door}           & Conservative Factor $\beta$                                                                               & 0.9                               \\
                                               & Minimum Trajectory Length\footref{refnote}                                                                & 0.75                              \\
                                               & Trajectories Proportion $\zeta$                                                                           & 1.0                               \\
                                               & Success Critic Network                                                                                    & MLP(256, 256)                     \\
                                               & Success Critic Optimizer                                                                                  & Adam                              \\
                                               & Success Critic LR                                                                                         & 3e-4                              \\
                                               & Success Critic Output Activation                                                                          & $-0.5cos(x) - 1$                  \\
                                               & \# Actions Sampled                                                                                        & 5                                 \\
        \midrule
        \multirow{7}{*}{Sawyer Peg}            & Conservative Factor $\beta$                                                                               & 0.9                               \\
                                               & Minimum Trajectory Length\footref{refnote}                                                                & 0.5                               \\
                                               & Trajectories Proportion $\zeta$                                                                           & 0.5                               \\
                                               & Success Critic Network                                                                                    & MLP(256, 256)                     \\
                                               & Success Critic Optimizer                                                                                  & Adam                              \\
                                               & Success Critic LR                                                                                         & 3e-4                              \\
                                               & Success Critic Output Activation                                                                          & $-0.5cos(x) - 1$                  \\
                                               & \# Actions Sampled                                                                                        & 5                                 \\
        \midrule
        \multirow{7}{*}{Minitaur}              & Conservative Factor $\beta$                                                                               & 0.95                              \\
                                               & Minimum Trajectory Length\footref{refnote}                                                                & 0.25                              \\
                                               & Trajectories Proportion $\zeta$                                                                           & 0.25                              \\
                                               & Success Critic Network                                                                                    & MLP(256, 256)                     \\
                                               & Success Critic Optimizer                                                                                  & Adam                              \\
                                               & Success Critic LR                                                                                         & 3e-4                              \\
                                               & Success Critic Output Activation                                                                          & $-0.5cos(x) - 1$                  \\
                                               & \# Actions Sampled                                                                                        & 5                                 \\
        \midrule
        \multirow{7}{*}{4 Rooms}               & Conservative Factor $\beta$                                                                               & 0.95                              \\
                                               & Minimum Trajectory Length\footref{refnote}                                                                & 0                                 \\
                                               & Trajectories Proportion $\zeta$                                                                           & 0.5                               \\
                                               & Success Critic Network                                                                                    & \makecell[l]{Conv([16,16,16], 3), \\FC(64)} \\
                                               & Success Critic Optimizer                                                                                  & Adam                              \\
                                               & Success Critic Output Activation                                                                          & $sigmoid(x)$                      \\
                                               & Success Critic LR                                                                                         & 1e-3                              \\
        \bottomrule
    \end{tabular}
    \caption{Hyperparameters for \method{}}
    \label{tab:sac_esfb_hyperparameters}
\end{table}
\section{Experimental Setup}
\label{sec:experimental_setup}
All experiments were run as CPU jobs. Table \ref{tab:env_details} gives further details about the experimental
setup for each environment. Each configuration is run for 5 seeds. Hyperparameter search
was done over the following hyperparameters: conservative factor $\beta\in\{0.0,
    0.9,0.95\}$, minimum trajectory length as a fraction of the maximum trajectory length
$\in\{0.0, 0.25, 0.5, 0.75\}$, and switching trajectories proportion $\zeta\in\{0.25, 0.5, 0.75,
    1.0\}$.
\begin{table}
    \small
    \centering
    \begin{tabular}{lllllll}
        \toprule
                    & \makecell[l]{\textbf{\# Train}                                          \\\textbf{Steps}} & \makecell[l]{\textbf{\# Train}                          \\ \textbf{Hours}} & \makecell[l]{\textbf{Hard Reset}         \\\textbf{Frequency}} & \makecell[l]{\textbf{Epsisode}         \\\textbf{length limit}} & \makecell[l]{\textbf{\# of Demo}\\\textbf{Transitions}} & \makecell[l]{\textbf{Reward}\\\textbf{Type}}\\
        \midrule
        \makecell[l]{Tabletop                                                                 \\Manipulation} & 3,000,000                      & 24 & 200,000 & 200   & 2,534 & Sparse \\
        Sawyer Door & 5,000,000                      & 42  & 200,000 & 300   & 1,095 & Sparse \\
        Sawyer Peg  & 7,000,000                      & 68  & 100,000 & 200   & 1,815 & Sparse \\
        Minitaur    & 5,000,000                      & 50  & 100,000 & 1,000 & 0     & Dense  \\
        4-Rooms     & 50,000                         & 0.5 & 50,000  & 100   & 0     & Sparse \\
        \bottomrule
    \end{tabular}
    \caption{Experimental setup details for each environment.}
    \label{tab:env_details}
\end{table}


\section{Metrics Computation}
The metrics in Figures \ref{fig:main_rliable} and \ref{fig:ablation_rliable} were all
calculated using the rliable library \citep{agarwal2021deep}. Each metric was calculated
using $2000$ bootstrap replications over normalized and aggregated results from all tasks.
For \minitaur{}, we did not have the run data for all of the baselines, only the final
mean and standard error. To overcome this and integrate \minitaur{} results in our
comparisons, we generated $1000$ sets of samples from the
distribution represented by the mean and standard error of each baseline, computed rliable metrics with
each set, and used the set with the highest mean, to give as much benefit of the doubt
as possible to the baselines. Despite this bias, the difference between the best set and
the worst set was quite small, with the largest difference being approximately $.05$ for
FBRL.

\section{EARL Results}
We present the results for the best policy of each method on the environments in the EARL benchmark
for direct comparison with previous methods in Table \ref{tab:earl_results}. Note, \method{} has the best performance
of reset free methods on 3 of 4 environments, and performs somewhat competitively on the
last environment.
\begin{table}
    \small
    \centering
    \begin{tabular}{lcccc}
        \toprule
                                  & \makecell{\textbf{Tabletop}                                                                                 \\\textbf{Organization}} & \makecell{\textbf{Sawyer}\\ \textbf{Door}} & \makecell{\textbf{Sawyer}\\ \textbf{Peg}} & \makecell{\textbf{Minitaur}} \\
        \midrule
        \textit{Naive RL}         & $0.32 \pm 0.17$             & $0.00 \pm 0.00$          & $0.00 \pm 0.00$            & $-1041.10 \pm 44.58$  \\
        \textit{R3L}              & $0.96 \pm 0.04$             & $0.54 \pm 0.18$          & $0.00 \pm 0.00$            & $-186.30 \pm 34.79$   \\
        \textit{VaPRL}            & $0.98 \pm 0.02$             & $0.94 \pm 0.05$          & $0.02 \pm 0.02$            & -                     \\
        \textit{MEDAL}            & $0.98 \pm 0.02$             & $\mathbf{1.00 \pm 0.00}$ & $\mathbf{0.40 \pm 0.16}$   & -                     \\
        \textit{FBRL}             & $0.94 \pm 0.04$             & $\mathbf{1.00 \pm 0.00}$ & $0.00 \pm 0.00$            & $-986.34 \pm 67.95$   \\
        \textit{\method{} (ours)} & $\mathbf{1.0  \pm  0.0}$    & $\mathbf{1.0  \pm 0.0}$  & $\mathbf{0.42  \pm  0.07}$ & $-321.74  \pm  53.73$ \\
        \midrule
        \textit{Oracle RL}        & $0.80 \pm 0.11$             & $1.00 \pm 0.00$          & $1.00 \pm 0.00$            & $-41.50 \pm 3.40$     \\
        \bottomrule
    \end{tabular}
    \caption{Average return of the best policy over 5 random seeds, reported with standard error.
        Higher returns are better, and the best method for each environment (excluding oracle) is in bold.
        For \tabletop{}, \sawyerdoor{}, and \sawyerpeg{}, minimum performance is 0.0 and maximum performance
        is 1.0.}
    \label{tab:earl_results}
\end{table}

\begin{figure}
    \centering
    \includegraphics[width=.7\textwidth]{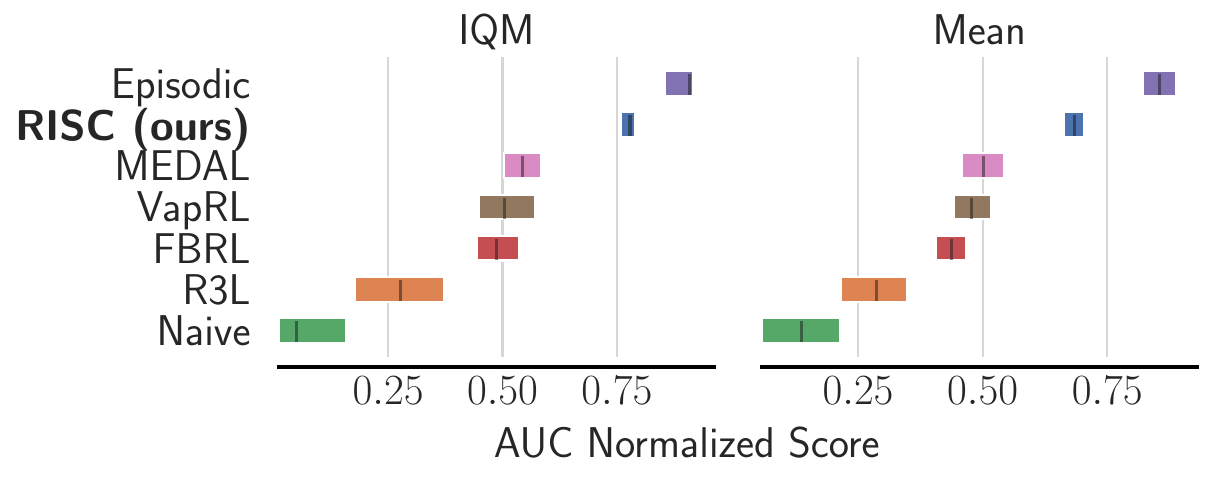}
    \caption{\new{Interquartile mean and mean for the area under the curve (AUC) for the
            learning curve of each run, aggregated across runs and environments. Higher AUC is
            better and implies more sample efficient learning.}
    }
    \label{fig:auc_rliable}
\end{figure}

\new{We also present the interquartile mean and mean for the area under the curve (AUC) of
    the learning curves for each agent, aggregated across seeds and environments (Figure
    \ref{fig:auc_rliable}). \method{}
    is significantly higher than the other methods, implying that it is more sample
    efficient.}

\section{Exploring the Modulating Mechanisms}
\begin{figure}[htbp]
    \centering
    \begin{subfigure}[b]{0.45\textwidth}
        \includegraphics[width=\textwidth]{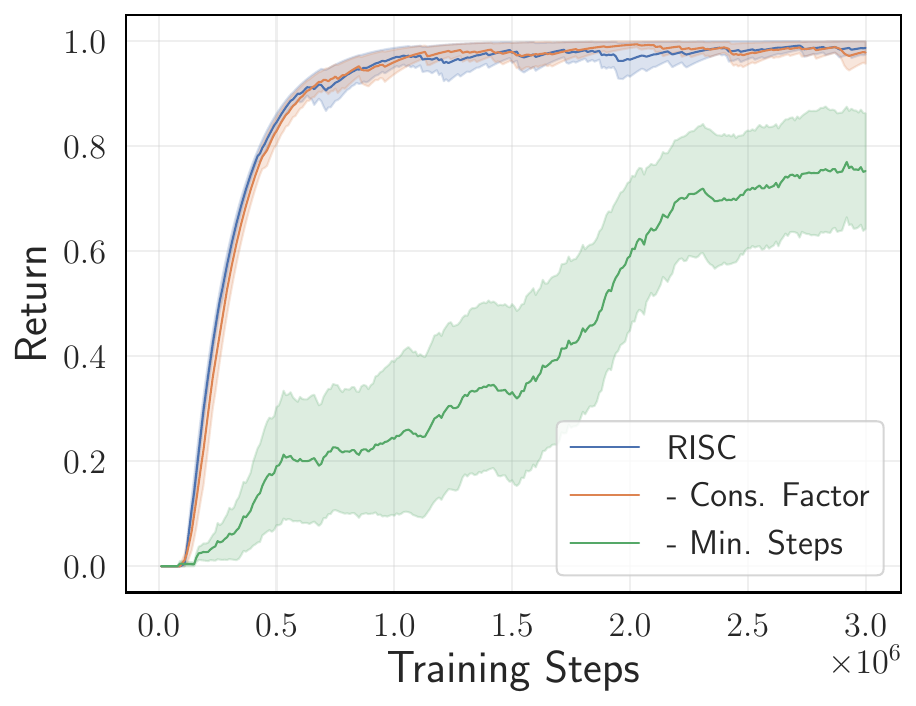}
        \caption{Tabletop Manipulation}
        \label{fig:tm_mod}
    \end{subfigure}
    \begin{subfigure}[b]{0.45\textwidth}
        \includegraphics[width=\textwidth]{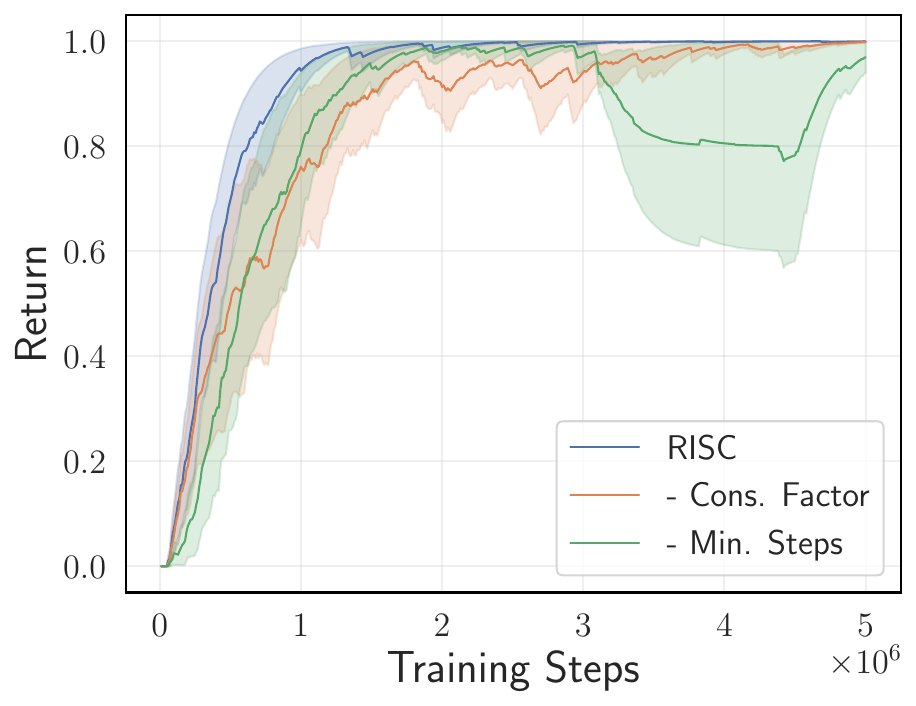}
        \caption{Sawyer Door}
        \label{fig:sd_mod}
    \end{subfigure}
    \begin{subfigure}[b]{0.45\textwidth}
        \includegraphics[width=\textwidth]{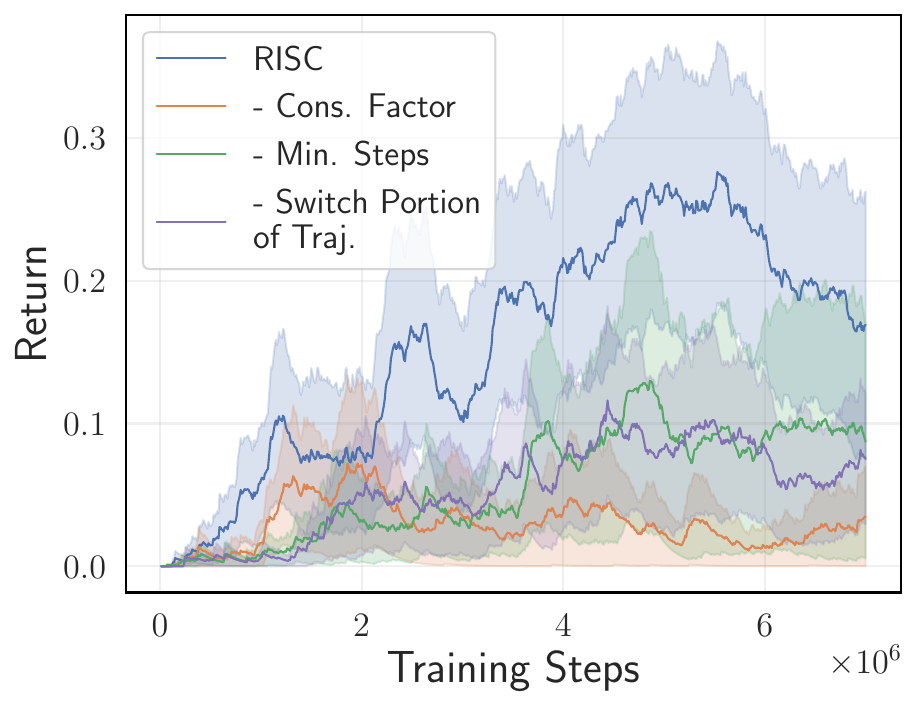}
        \caption{Sawyer Peg}
        \label{fig:sp_mod}
    \end{subfigure}
    \begin{subfigure}[b]{0.45\textwidth}
        \includegraphics[width=\textwidth]{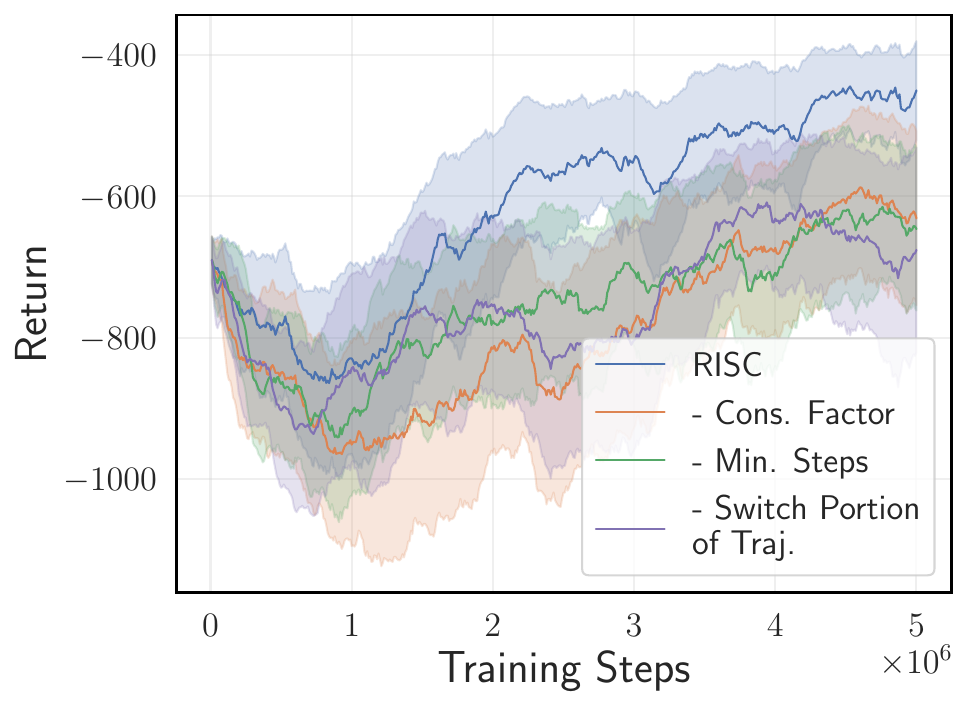}
        \caption{Minitaur}
        \label{fig:minitaur_mod}
    \end{subfigure}
    \caption{We show the effect of removing the modulations presented in Section
        \ref{subsec:modulations}. The performance of the agent degrades across all
        environments, with significant deterioration on \minitaur{} and \sawyerpeg{}.}
    \label{fig:mod_results}
\end{figure}

\begin{figure}[htbp]
    \centering
    \begin{subfigure}[b]{0.45\textwidth}
        \includegraphics[width=\textwidth]{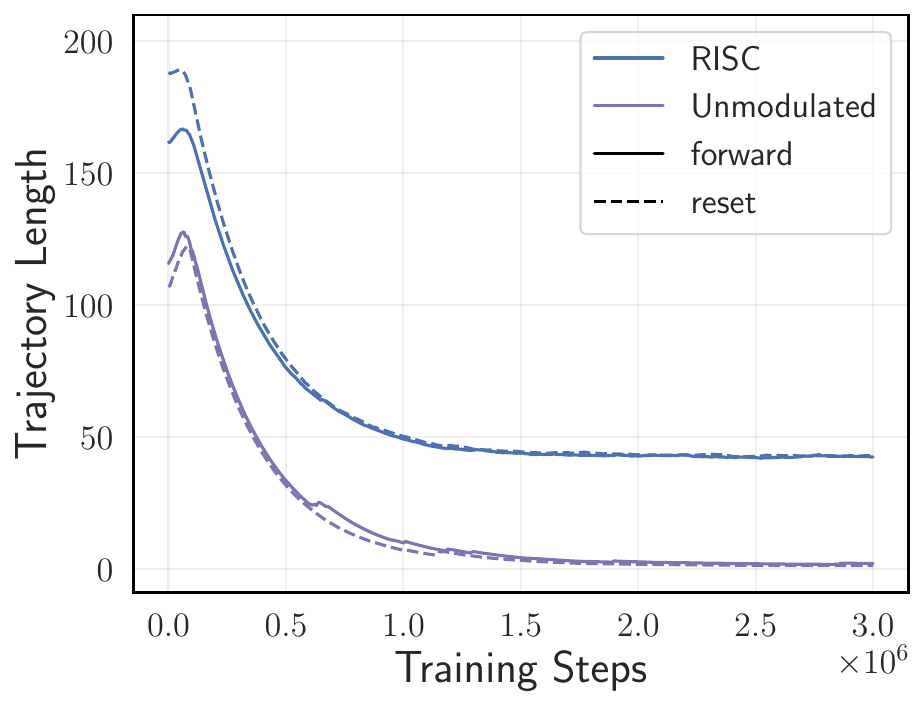}
        \caption{Tabletop Manipulation}
        \label{fig:tm_tl}
    \end{subfigure}
    \begin{subfigure}[b]{0.45\textwidth}
        \includegraphics[width=\textwidth]{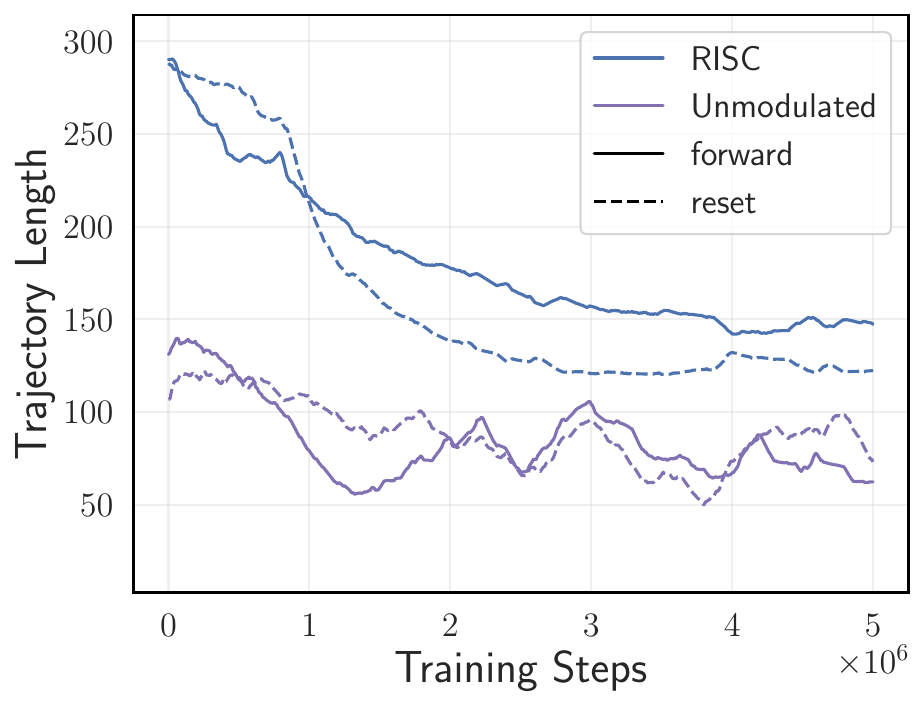}
        \caption{Sawyer Door}
        \label{fig:sd_tl}
    \end{subfigure}
    \begin{subfigure}[b]{0.45\textwidth}
        \includegraphics[width=\textwidth]{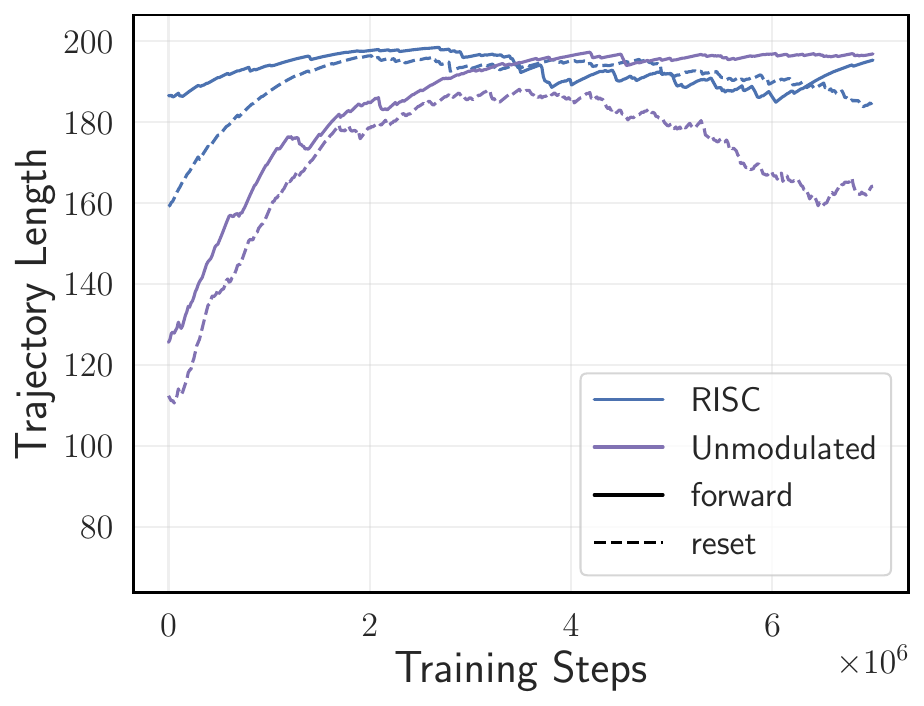}
        \caption{Sawyer Peg}
        \label{fig:sp_tl}
    \end{subfigure}
    \begin{subfigure}[b]{0.45\textwidth}
        \includegraphics[width=\textwidth]{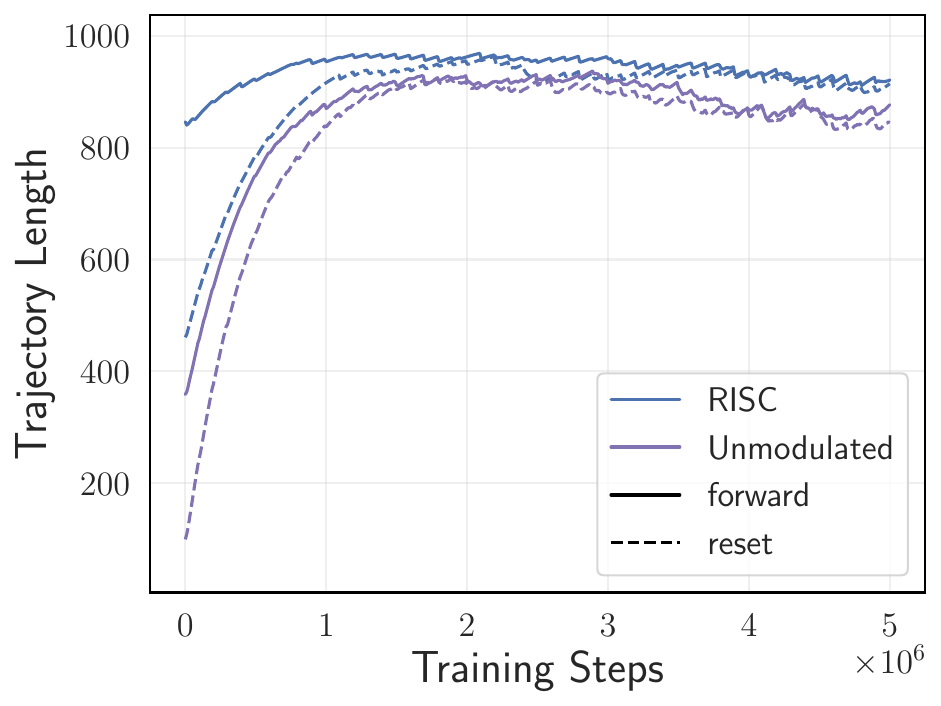}
        \caption{Minitaur}
        \label{fig:minitaur_tl}
    \end{subfigure}
    \caption{We show the effect of removing the modulations presented in Section
        \ref{subsec:modulations} on the trajectory lengths of the agent. As expected,
        the average trajectory length decreases significantly on all environments.}
    \label{fig:mod_tl}
\end{figure}

In this section, we show the necessity of the modulations introduced in Section
\ref{subsec:modulations}.
In Figure \ref{fig:mod_tl}, we see that removing the modulations significantly reduces
the average trajectory length of the agent.
Furthermore, when we compare the performance of the \method{} with and
without modulations in Figure
\ref{fig:mod_results}, we see that without the modulations, \method{}
underperforms on all environments.

\section{\new{Hyperparameter Analysis}}

\begin{figure}
    \centering
    \includegraphics[width=\textwidth]{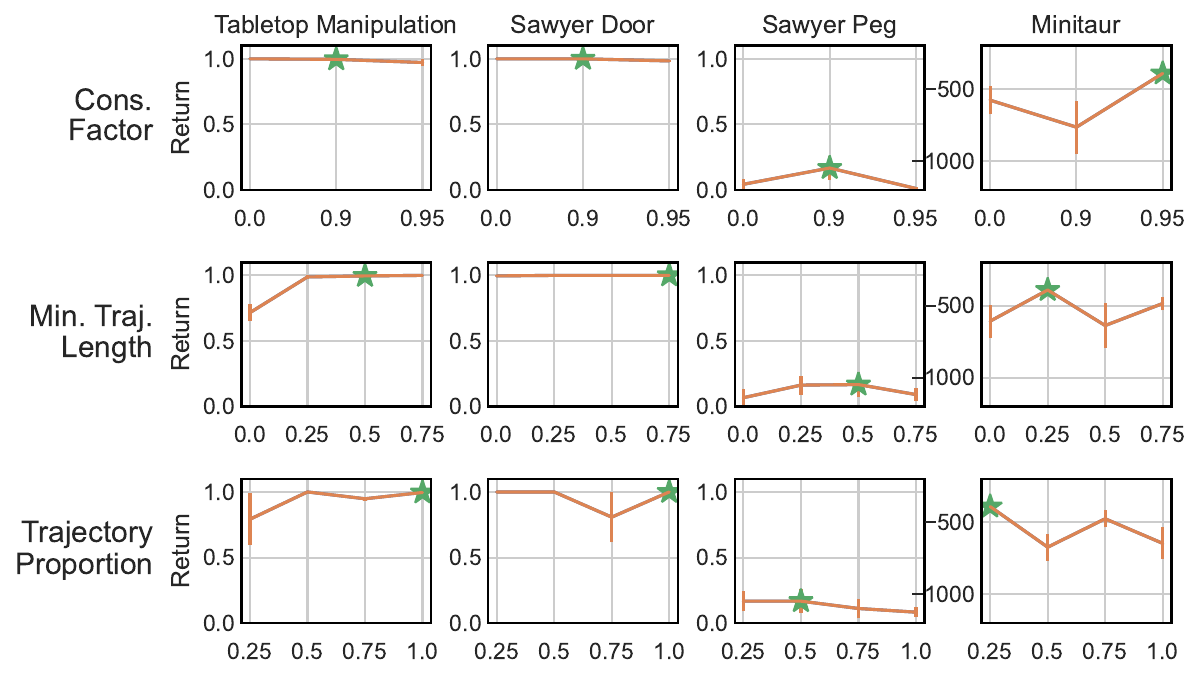}
    \caption{\new{Hyperparameter sensitivity analysis. We vary the selected configuration one
            hyperparameter at a time, and present the mean and standard deviation for each
            configuration. The configurations used in the experiments are marked with a green star.}
    }
    \label{fig:hparam_sensitivity}
\end{figure}

\new{We first present details of our
    hyperparameter sensitivity analysis.
    We took the optimal configurations selected for each environment, and varied the value
    of one hyperparameter at a time.
    The results in Figure \ref{fig:hparam_sensitivity} show that our results are fairly
    robust to changes in hyperparameters, with
    fairly small changes to the final return across all environments.}

\begin{figure}
    \centering
    \begin{subfigure}[b]{\textwidth}
        \includegraphics[width=\textwidth]{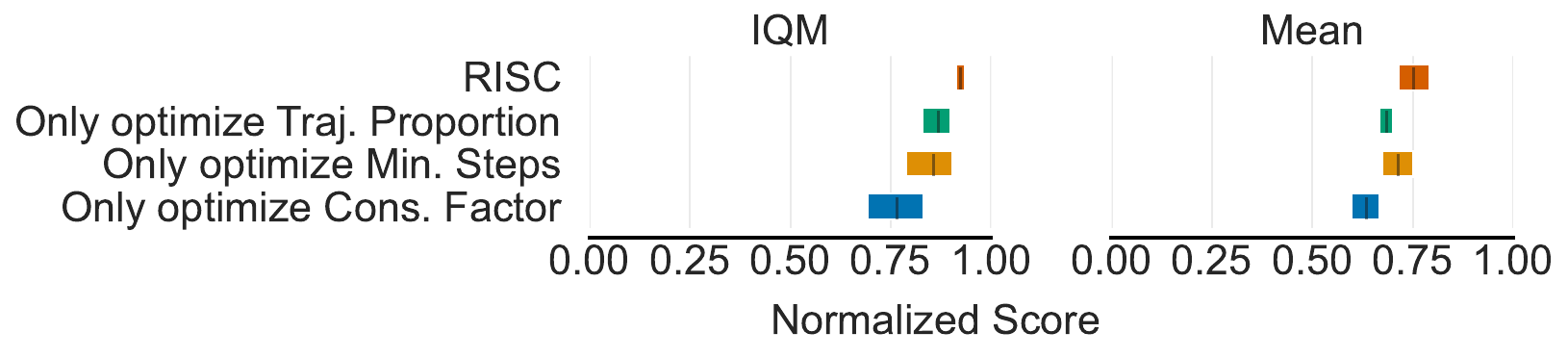}
        \caption{\new{Results when only optimizing 1 hyperparameter at a time.}}
        \label{fig:opt_1}
    \end{subfigure}
    \begin{subfigure}[b]{\textwidth}
        \includegraphics[width=\textwidth]{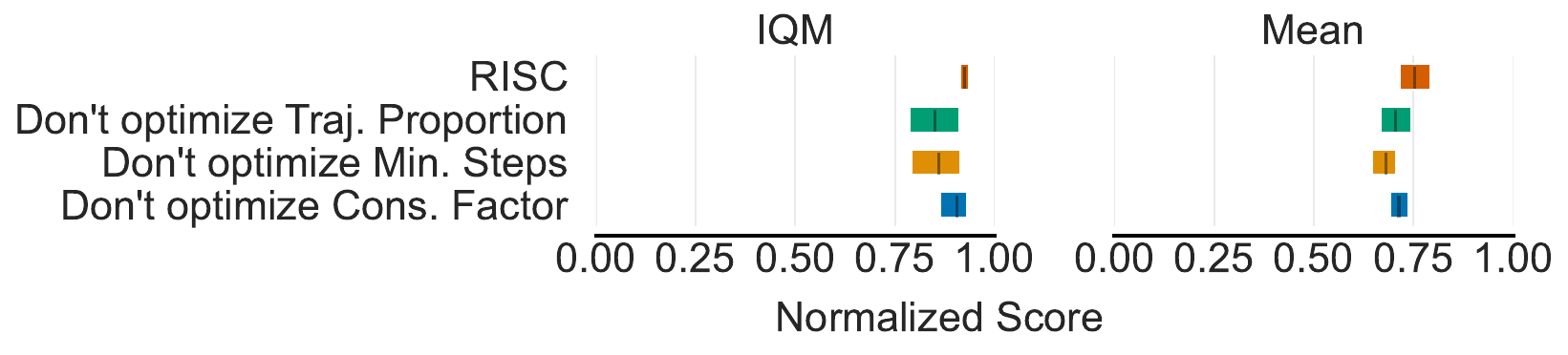}
        \caption{\new{Results when only optimizing 2 hyperparameters at a time.}}
        \label{fig:opt_2}
    \end{subfigure}
    \caption{\new{Results of partial hyperparameter optimization.}}
\end{figure}

\new{For our second analysis, we tried optimizing only 1 hyperparameter at a time.
    For each environment, we select the configuration with the best performance for each
    hyperparameter where the other modulations are turned off, and aggregate across
    environments.
    We do a similar analysis with optimizing 2 hyperparameters at a time. The results are
    shown in Figure \ref{fig:opt_1} and Figure \ref{fig:opt_2} respectively.
    The most important hyperparameters seem to be the minimum trajectory length or the
    proportion of trajectories where the switching happens. They recover most of the
    performance, but the full method still results in a slight performance gain.}

